%% file: iclr2025_conference.tex
\documentclass{article} 
\usepackage{iclr2025_conference,times}

\input{math_commands.tex}

\usepackage{hyperref}
\usepackage{url}

\usepackage{graphicx} 

\newcommand{\PDE}{\mathrm{PDE}}  
\newcommand{\IC}{\mathrm{IC}}
\newcommand{\BC}{\mathrm{BC}}
\renewcommand{\R}{\mathbb{R}}  
\renewcommand{\L}{\mathcal{L}}  
\newcommand{\FourierEmb}{\gamma}

\newcommand{\NN}{\mathrm{NN}}  
\newcommand{\trafo}{\mathrm{trafo}} 

\usepackage{multirow}
\usepackage{makecell}

\title{Hard-constraining Neumann boundary conditions in physics-informed neural networks via Fourier feature embeddings}

\author{Christopher Straub, Philipp Brendel, Vlad Medvedev \& Andreas Rosskopf \\
	Department Modeling and Artificial Intelligence\\
	Fraunhofer Institute for Integrated Systems and Device Technology IISB\\
	Schottkystrasse 10, 91058 Erlangen, Germany \\
	\texttt{\{christopher.straub,philipp.brendel,vlad.medvedev,}\\
	\texttt{\quad andreas.rosskopf\}@iisb.fraunhofer.de}
}

\iclrfinalcopy 
\begin{document}

\maketitle

\begin{abstract}
	We present a novel approach to hard-constrain Neumann boundary conditions in physics-informed neural networks (PINNs) using Fourier feature embeddings. 
	Neumann boundary conditions are used to described critical processes in various application, yet they are more challenging to hard-constrain in PINNs than Dirichlet conditions.
	Our method employs specific Fourier feature embeddings to directly incorporate Neumann boundary conditions into the neural network's architecture instead of learning them.
	The embedding can be naturally extended by high frequency modes to better capture high frequency phenomena.
	We demonstrate the efficacy of our approach through experiments on a diffusion problem, for which our method outperforms existing hard-constraining methods and classical PINNs, particularly in multiscale and high frequency scenarios.
\end{abstract}

\section{Introduction}\label{sc:intro}

Diffusion equations are essential in the study of multiscale processes such as semiconductor modeling~\citep{Aleksandrov}, battery modeling~\citep{Battery_appl}, biological systems~\citep{Biology_appl}, and larger scale phenomena like atmospheric and oceanic transport processes~\citep{Atmo_appl,Ocean_appl}.
These problems often contain Neumann boundary conditions, specifying the gradient or flux at a boundary, to represent symmetries or essential physical processes like insulation, recharge rates, or surface charge densities.
Physics-informed neural networks (PINNs) offer a promising alternative to classical numerical solvers to simulate such systems~\citep{PINN19}.
The main idea of PINNs is to represent the unknown solution of a physical system as a neural network and train it using the governing physical laws.
Despite PINNs being successfully applied in a variety of different settings~\citep{PIML24,PINN_appl_24}, they face some difficulties.
As argued in~\citet{Wang21,Wang22}, two major challenges faced by PINNs are balancing multiple learning objectives (learning one or more differential equations and boundary conditions) and accurately approximating high frequency or multiscale phenomena.

One way to address the former challenge is to incorporate certain requirements on the solution, e.g., boundary conditions, directly into the neural network's structure instead of learning them.
This is known as {\em hard-constraining} and is typically achieved by adding carefully chosen, non-trainable layers to the start or the end of a neural network to transform its input or output, respectively.
Hard-constraining techniques are commonly used for Dirichlet or periodic boundary conditions~\citep{PIML24}.
Due to the simplification of the learning task, they typically improve the performances of PINNs~\citep{hPINN21,SuSr22,Zeinhofer24}.
Neumann boundary conditions are more challenging to hard-constrain and consequently less frequently hard-constrained, see \secref{ssc:HCNeumann} for a review of existing methods in this direction.

To allow neural networks to approximate high frequency or multiscale data, it has been proposed in~\cite{Fourier20} to expand the input of the neural network into Fourier modes of different frequencies.
In the seminal work on the application of PINNs to multiscale problems~\cite{Wang21}, it has been shown theoretically and experimentally that such {\em Fourier feature embeddings} help PINNs to capture high frequency and multiscale phenomena.

In this paper, we present a new way of hard-constraining Neumann boundary conditions based on carefully chosen Fourier feature mappings. 
Concretely, our contributions are:
\begin{itemize}
	\item We present a new method to hard-constrain Neumann boundary conditions into PINNs by suitably transforming their input using a single, scalar Fourier embedding.
	This embedding can further be extended by higher frequencies to better capture high frequency phenomena.
	\item We demonstrate the efficacy of the new method for a forward diffusion problem with Neumann boundary conditions. 
	The new method yields more accurate results than existing PINN methods, with the most significant improvements observed for high frequencies and for differing frequencies on different scales.
\end{itemize}

Despite the fact that the numerical experiments conducted in the present workshop paper are limited to a one-dimensional forward problem, it should be emphasized that the method is highly versatile and can also be applied in various other contexts. 
A selection of these is outlined in appendix~\ref{sc:metho_general}, including the case of a higher dimensional spatial domain.


\section{Theoretical Background}\label{sc:background}

In this section we provide the necessary backgrounds on which the new method relies.
As an explicit example to introduce the methods, we consider the one-dimensional heat equation with spatial and temporal domains both normalized to unity, i.e.,
\begin{equation}\label{eq:heat}
	\partial_tu=D\,\partial_x^2u,\qquad(x,t)\in[0,1]^2,
\end{equation}
where $D>0$ is a diffusivity parameter.
This equation is coupled with vanishing Neumann boundary conditions
\begin{equation}\label{eq:Neumann_heat}
	\partial_xu(0,t)=0=\partial_xu(1,t),\qquad t\in[0,1],
\end{equation}
and initial condition
\begin{equation}\label{eq:initial_heat}
	u(x,0)=g(x),\qquad x\in[0,1],
\end{equation}
for prescribed $g\colon[0,1]\to\R$.
Theoretical background on this system is given in appendix~\ref{sc:heat}.

\subsection{Physics-informed neural networks}\label{ssc:PINNs}

The strategy of PINNs is to represent the sought solution of a physical system as a neural network and train it using the describing physical laws~\citep{PINN19}.
The neural network's architecture is typically a multilayer perceptron (MLP), although all of the below also applies to different architectures, e.g., a Kolmogorov-Arnold network~\citep{KAN24,PIKAN24}.
The parameters of the neural network are trained by minimizing the composite loss function
$\L=\lambda_\PDE\,\L_\PDE + \lambda_\IC\,\L_\IC + \lambda_\BC\,\L_\BC.$
The individual loss terms $\L_\PDE$, $\L_\BC$, and $\L_\IC$ are (usually) the mean squared error of the residual of the PDE, the initial condition, and the boundary condition evaluated at randomly sampled collocation points $(x_j^\PDE,t_j^\PDE)_{j=1,\ldots,N_\PDE}\subset[0,1]^2$, $(x_j^\IC,0)_{j=1,\ldots,N_\IC}\subset[0,1]\times\{0\}$, and $(x_j^\BC,t_j^\BC)_{j=1,\ldots,N_\BC}\subset\{0,1\}\times[0,1]$, respectively.
The weights $\lambda_\PDE, \lambda_\IC, \lambda_\BC>0$ can be used to balance the individual loss contributions. Here, the default choice $\lambda_\PDE=\lambda_\IC=\lambda_\BC=1$ is used.
More background on PINNs can be found in~\citet{PINN19,PINN98,PIML21,Experts23}.

\subsection{Existing hard-constraining techniques for Neumann boundary conditions}\label{ssc:HCNeumann}


The main idea of hard-constraining is to incorporate certain conditions on the sought solution directly into the neural network's structure.
An approach to achieve this for Neumann boundary conditions on general domains has been presented in~\citet{SuSr22}.
It is based on distance functions, i.e., functions giving the signed distance to the boundary of the problem's domain.
These distance functions are then used to hard-constrain different types of boundary conditions into neural networks. 
The formulae to hard-constrain Neumann boundary conditions are given \mbox{in~\citet[Sc.~5.1.2]{SuSr22}} and rely on a suitable transformation of the neural network's output to modify the value of its derivatives at the boundary.
To explicitly demonstrate this method in the case of the vanishing Neumann boundary conditions given by~\eqref{eq:Neumann_heat}, let $u^\NN\colon[0,1]^2\to\R$ be an arbitrary smooth representation model, e.g., a neural network.
Inserting the canonical distance function $\phi(x)=x(1-x)$ of~$[0,1]$ into the formulae from \citet{SuSr22} yields the following transformation of~$u^\NN$:
\begin{equation}\label{eq:utrafo_lit}
	u^\trafo(x,t)=u^\NN(x,t)-x(1-x)^2\,\partial_xu^\NN(0,t)-x^2(x-1)\,\partial_xu^\NN(1,t).
\end{equation}
The original output~$u^\NN$ is replaced by this transformed version~$u^\trafo$ when evaluating the model and when computing the loss~$\L$.
The key idea of this transformation is to use the values of the $x$-derivative of the original representation model~$\partial_xu^\NN$ to ensure the desired behavior of the transformation~$u^\trafo$ at the spatial boundary. 
Indeed, a straight-forward calculation shows that~$u^\trafo$ satisfies the vanishing boundary conditions independently of $u^\NN(x,t)$, and hence the loss term~$\L_\BC$ in~$\L$ can be dropped when training the model. 
In~\cite{SuSr22}, it is observed that this remains true if the additional term $x^2(1-x)^2v^\NN(x,t)$ is added to the right-hand side of~\eqref{eq:utrafo_lit}, where $v^\NN=v^\NN(x,t)$ is another trainable representation model.
Note, however, that the computational costs of evaluating or backpropagating the transformed model~$u^\trafo$ are significantly increased by incorporating the additional derivatives~$\partial_xu^\NN$ in~\eqref{eq:utrafo_lit}, cf.\ \secref{sc:experiments} for details.

The similar method in the case of an ordinary differential equation with a single Neumann condition has also been proposed in~\cite{HC_Neumann20}.
Mathematically more advanced schemes in the situation where Neumann boundary conditions are imposed together with Dirichlet conditions are presented in~\cite{HC_Neumann09,HC_Neumann09-2}. The latter also rely on output transformations containing suitable evaluations of the neural network at the boundary.

\subsection{Fourier feature embeddings}\label{ssc:Fourier}

The main idea of Fourier feature embeddings is to expand the input variable(s) of a neural network into multiple frequencies to help the model approximate phenomena with different frequencies, in particular, high frequencies.
This has first been proposed in~\cite{Fourier20} in the context of computer vision problems.
In~\cite{Wang21}, it has then been demonstrated that adding Fourier feature embeddings to PINNs allow them to learn multiscale processes while vanilla PINNs struggle at this task.
%
Concretely, for the diffusion problem given by \eqrefsys{eq:heat}{eq:initial_heat}, the Fourier feature embedding of the spatial coordinate $x\in[0,1]$ proposed in~\cite{Wang21} is
\begin{equation*}
	x\mapsto(\cos(b_1\pi x),\sin(b_1\pi x),\ldots,\cos(b_m\pi x),\sin(b_m\pi x)),
\end{equation*}
for frequencies $b_1,\ldots,b_m\in\R$.
Together with the time variable~$t$, this spatial embedding is then input into a trainable neural network.
In~\cite{Wang21}, it is further shown how to use multiple spatial Fourier embeddings as well as additional temporal Fourier embeddings. 
In the present work, we restrict the discussion to the use of one spatial Fourier embedding of a prescribed length~$2m$. 
The frequencies used for the Fourier embedding are typically sampled from a Gaussian distribution with zero mean and prescribed standard deviation~$\sigma$ -- this is referred to as a {\em random} Fourier feature embedding.
Other approaches are to learn the frequencies~\citep{DoNi21} or to explicitly prescribe them.
%
Fourier feature embeddings can also be used to hard-constrain periodic boundary conditions into PINNs~\citep{DoNi21,hPINN21}.
This requires the frequency values $b_1,\ldots,b_m$ to be chosen such that the Fourier mappings are periodic on the problem's domain.

\section{Methodology}\label{sc:metho}

\begin{figure}[h!]
	\begin{center}
		\includegraphics[width=0.9\columnwidth]{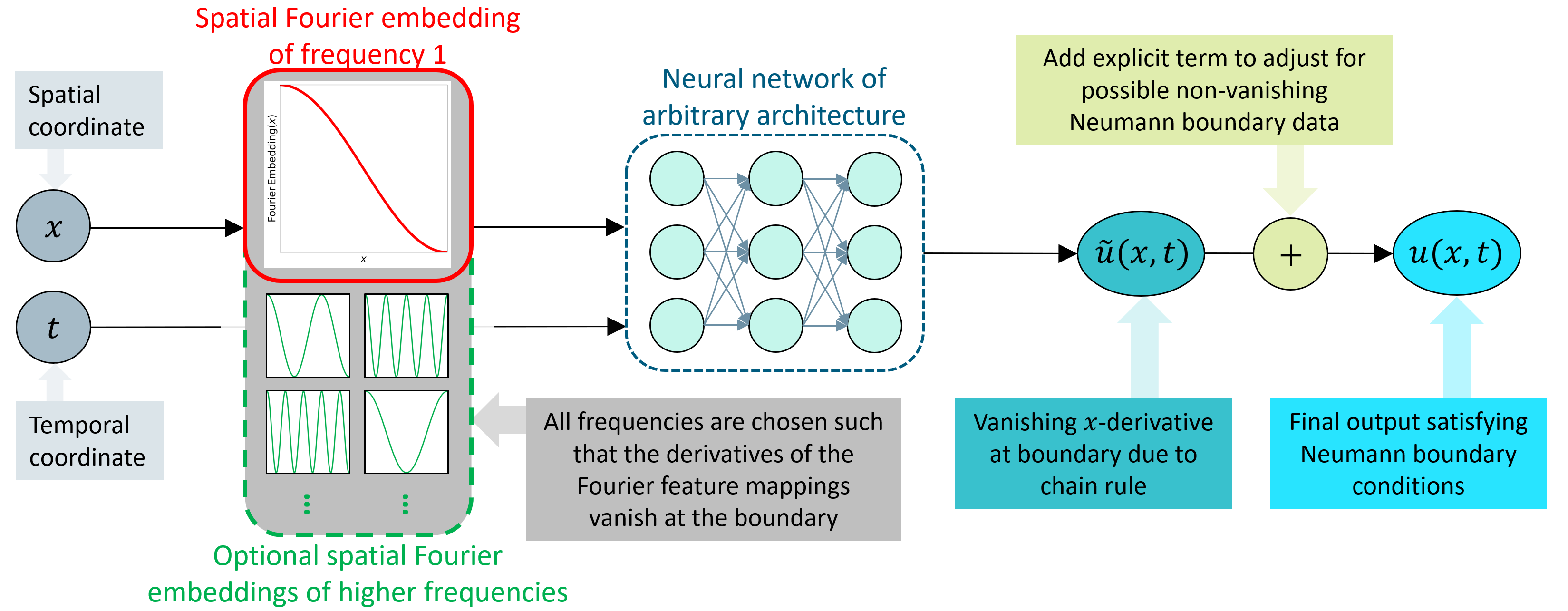}
	\end{center}
	\caption{The proposed method to hard-constrain Neumann boundary conditions.}
	\label{fig:method}
\end{figure}

We now present a new approach to hard-constrain Neumann boundary conditions in PINNs.
To introduce the method, we focus on the vanishing Neumann boundary conditions from \eqref{eq:Neumann_heat} imposed on the spatial domain $x\in[0,1]$.
Nonetheless, the method is straight-forward to extend to non-vanishing Neumann boundary conditions and to more general and higher dimensional spatial domains, see appendix~\ref{sc:metho_general} for the formulae in these more general settings.
The key mechanisms of the new method are illustrated in \figref{fig:method}:
A Fourier embedding of frequency~$1$ (red) is applied to the spatial input~$x$. The Fourier embedding can naturally be extended by higher frequency embeddings (green). 
The key idea is to only use specific Fourier mappings (only the Cosine with integer frequencies) that are flat at the spatial boundary.
Together with additional input coordinates (e.g., the time~$t$), the whole Fourier embedding is input into a trainable neural network. Due to the chain rule, the derivative w.r.t.\ the spatial variable~$x$ vanishes at the boundary of the spatial domain.
To adjust for non-vanishing Neumann boundary data, an explicit expression is added to the neural network's output to arrive at the model's final output.

Let us now provide the mathematical formulae underpinning these main ideas.
The single spatial Cosine Fourier embedding of frequency~$1$ is given by the following transformation to the spatial input variable $x\in[0,1]$:
\begin{equation}\label{eq:FourierEmb_NewSingle}
	x\mapsto\FourierEmb_1(x)=\cos(\pi x).
\end{equation}
The key property of $\FourierEmb_1$ is that it is flat at the spatial boundary, i.e., $\FourierEmb_1'(x)=0$ for $x=0$ and $x=1$. 
This is also true for the higher frequency embedding $x\mapsto\cos(\pi bx)$ 
provided that~$b$ is an integer.
To better capture high frequency phenomena, these embeddings can be appended to~$\FourierEmb_1(x)$ to arrive at the total embedding
\begin{equation}\label{eq:FourierEmb_NewMult}
	x\mapsto\FourierEmb_{1,b_2,\ldots,b_{n}}(x)=\left(\cos(\pi x), \cos(\pi b_2x),\ldots,\cos(\pi b_{n}x)\right),
\end{equation}
where the integer frequencies $b_2,\ldots,b_n\in\mathbb Z$ are randomly sampled.
The embedding of frequency~$1$ is always included in \eqref{eq:FourierEmb_NewMult} to ensure that no information is lost in the transformation -- note that $\FourierEmb_1\colon[0,1]\to[-1,1]$ is one-to-one and $\FourierEmb_1'(x)\neq0$ for $0<x<1$.
The embedding given by~\eqref{eq:FourierEmb_NewSingle} or~\eqref{eq:FourierEmb_NewMult} is then inserted into an arbitrary smooth representation model~$u^\NN$, e.g., a neural network.
The resulting model is
\begin{equation}\label{eq:utrafo_our}
	\tilde u(x,t)=u^\NN(\FourierEmb(x),t),
\end{equation} 
where $\FourierEmb$ can be $\FourierEmb_1$ or $\FourierEmb_{1,b_2,\ldots,b_{n}}$.
Because the derivative of $\FourierEmb$ vanishes at the boundary, i.e., $\frac d{dx} \FourierEmb(x)=0$ for $x=0$ and $x=1$, the chain rule implies that~$\tilde u$ indeed satisfies the vanishing Neumann boundary condition from \eqref{eq:Neumann_heat}.
Similar to \secref{ssc:HCNeumann}, this property remains valid if the additional term $x^2(1-x)^2v^\NN(x,t)$ is added to the right-hand side of~\eqref{eq:utrafo_lit}, where $v^\NN=v^\NN(x,t)$ is another trainable representation model.
Hard-constraining non-vanishing Neumann boundary conditions, i.e., $\partial_xu(0,t)=A$ and $\partial_xu(1,t)=B$ for prescribed $A,B\in\R$ instead of~\eqref{eq:Neumann_heat}, is achieved by the transformation
\begin{equation}\label{eq:utrafo_our_nonvanishing}
	u(x,t)=u^\NN\big(\FourierEmb(x),t\big) + x(1-x)^2A+x^2(x-1)B.
\end{equation} 
Notice that this transformation is fundamentally simpler than the one presented in \secref{ssc:HCNeumann}: 
In~\eqref{eq:utrafo_lit}, derivatives of the neural network evaluated at the boundary are added in the transformation, while the functions added in~\eqref{eq:utrafo_our_nonvanishing} are explicit expressions of the spatial variable~$x$.

\section{Experiments}\label{sc:experiments}

In this section, the performance of the newly proposed method (cf.\ \secref{sc:metho}) is compared with the existing methods (cf.\ \secref{sc:background}) for the diffusion problem given by \eqrefsys{eq:heat}{eq:initial_heat}.
The same problem has been considered in~\cite{Wang21}, with the exception that Neumann instead of Dirichlet boundary conditions are used here.
The reason for considering this problem is that it exhibits a multitude of qualitatively different solutions for different initial data.
The initial conditions used here are given in \tabref{table:initials_heat}.
Low or high frequency initial data lead to the respective spatial frequencies of the solution with an additional temporal dependency.
In the multiscale case, the solution manifests low frequency on a macro scale and high frequency on a micro scale, resembling numerous practical multiscale scenarios~\citep{Wang21}.
As canonical examples for general initial conditions, polynomials of orders~$3$ and~$4$ satisfying the vanishing Neumann boundary conditions are considered.
These polynomials are strictly increasing and parabola-shaped, respectively, and are normalized to range in $[0,1]$.
They result in solutions composed of superpositions of infinite spatial frequencies, although the solutions are dominated by low frequency effects.
The diffusivity parameters~$D$ are chosen such that the solutions' temporal decay speeds are similar (cf.\ appendix~\ref{sc:heat}) -- 
the same strategy has been used in~\cite{Wang21}.

\begin{table}[h!]
	\caption{Initial conditions and diffusivity parameters in the diffusion problem~(\eqrefsys{eq:heat}{eq:initial_heat}) considered for the numerical experiments. The resulting solutions are stated in appendix~\ref{sc:heat}.}
	\label{table:initials_heat}
	\begin{center}
		\begin{tabular}{|c|c|c|} 
			Description & Initial condition & Diffusivity parameter \\
			\hline
			Low frequency & $g(x)=\cos(2\pi x)$ & $D=(2\pi)^{-2}$  \\
			High frequency & $g(x)=\cos(50\pi x)$ & $D=(50\pi)^{-2}$  \\
			Multiscale & $g(x)=\cos(2\pi x)+0.1\cos(50\pi x)$ & $D=(50\pi)^{-2}$   \\
			Polynom 3rd order & $g(x)=3x^2-2x^3$ & $D=\pi^{-2}$  \\
			Polynom 4th order & $g(x)=16x^4-32x^3+16x^2$ & $D=\pi^{-2}$  \\
		\end{tabular}
	\end{center}
\end{table}

The basic hyperparameters of all models are chosen similarly to~\cite{Wang21}, see appendix~\ref{sc:details_experiments} for details. 
The code is written in the widely used PINN framework DeepXDE~\citep{DeepXDE21}. 
Fourier embeddings are easy to implement in DeepXDE by using the pre-build function \texttt{apply\_feature\_transform}.
Implementing the existing hard-constraining technique (cf.\ \secref{ssc:HCNeumann}) is not that straightforward due to the derivative evaluation of the neural network contained in~\eqref{eq:utrafo_lit}. It can be realized by suitable modifications to the neural network class.

For each of the settings from \tabref{table:initials_heat}, nine different PINN methods are compared with one another.
Firstly, the vanilla PINN approach is considered, i.e., no hard-constraining of the Neumann boundary condition.
Secondly, a PINN with the existing approach to hard-constrain Neumann boundary conditions (recall \secref{ssc:HCNeumann}) is examined.
In addition, for both of these methods the use of random Fourier feature embeddings (recall \secref{ssc:Fourier}) is analyzed.
More precisely, Fourier embeddings of sizes~$20$ and~$50$ with random frequencies sampled from~$\mathcal N(0,20)$ are studied. 
This distribution and the embedding sizes have been chosen according to the guidelines outlined in~\citet{Wang21}.
Lastly, the newly proposed approach of hard-constraining Neumann boundary conditions is considered (recall \secref{sc:metho}).
To compare its performance with the aforementioned methods, a Fourier embedding with single frequency~$1$ (cf.\ \eqref{eq:FourierEmb_NewSingle}) as well as Fourier embeddings of sizes $20$ and~$50$ (cf.\ \eqref{eq:FourierEmb_NewMult}) with random (integer) frequencies sampled from~$\mathcal N(0,20)$ are used.
For the sake of clarity, we abstain from adding the term $x^2(1-x)^2v^\NN(x,t)$ with an additional trainable neural network $v^\NN$ to the output of both the existing and the newly proposed method of hard-constraining.

The performances of the different methods are compared in the same way as in~\cite{ConFIG} by computing the relative improvements of the accuracies w.r.t.\ a reference method. 
For instance, relative improvements of $+50\%$ and $-100\%$ correspond to half and twice the error of the reference method, respectively.
The reference method is always the vanilla PINN (i.e., no hard-constraining) with the Fourier embedding strategy leading to the best accuracy.
The accuracy always refers to the relative $L^2$-distance to the analytical solution. 
To obtain a fair comparison, all models are trained for the same time -- the time it took the vanilla PINN to complete~$10^6$ iterations.
The relative improvements of all methods are shown in \figref{fig:acc_bars}, the accuracies are stated in \tabref{table:experiments} in appendix~\ref{sc:details_experiments}.

\begin{figure}[h]
	\begin{center}
		\includegraphics[width=\columnwidth]{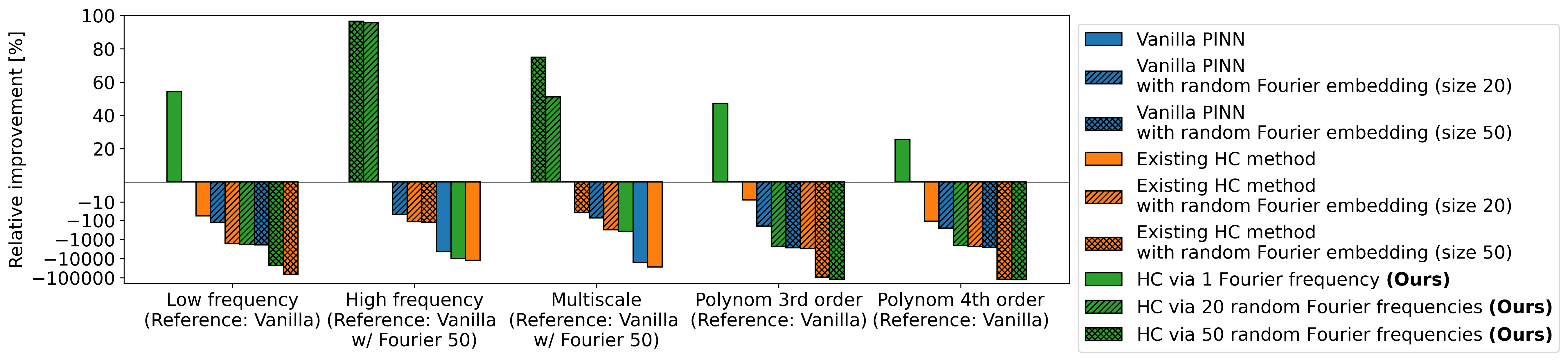}
	\end{center}
	\caption{Relative improvements of the accuracies of PINNs with different hard-constraining (HC) and Fourier embedding strategies w.r.t.\ the reference PINN. All models have been trained for the same time. More details on the simulations are provided in \figref{fig:acc_bars_allit} in appendix~\ref{sc:details_experiments}.}
	\label{fig:acc_bars}
\end{figure}

In the cases where the solution is dominated by low frequency effects, hard-constraining Neumann boundary conditions via a single Fourier embedding of frequency~$1$ yields the best accuracies.
Concretely, compared to the best vanilla PINN (no hard-constraining and no Fourier embeddings), this newly proposed method achieves relative improvements of $54.2\%$, $47.2\%$, and $25.6\%$ for low frequency, 3rd order polynomial, and 4th order polynomial initial conditions, respectively.
In all of these three cases, the existing hard-constraining technique performs slightly worse than the vanilla PINN.
Adding high frequency random Fourier embeddings results in a substantial deterioration in the performance of all hard-constraining techniques, where it can be observed that the accuracy decreases as the size of the Fourier embedding increases from~$20$ to~$50$.
The reason for this is that the PINN's training process to approximate a low frequency solution becomes significantly more complex and unstable when the spatial input~$x$ is mainly provided via high frequency feature embeddings.

For the high frequency and multiscale initial conditions, the reference method is the vanilla PINN with random Fourier embedding of size~$50$.
Better performances are achieved only by the newly proposed method of using Fourier embeddings of~$20$ or~$50$ random integer frequencies to hard-constrain Neumann boundary conditions.  
Concretely, hard-constraining the Neumann boundary conditions via a Fourier embedding of size~$50$ leads to a relative improvement of $96.5\%$ and $74.8\%$ in the high frequency and multiscale setting, respectively. 
The existing hard-constraining technique combined with a random Fourier embedding again performs slightly worse than the reference method.
It can further be seen in \figref{fig:acc_bars} that using smaller Fourier embeddings or no Fourier embedding leads to considerably worse accuracies (across all hard-constraining techniques), which is consistent with the analysis in~\cite{Wang21}.

In all settings, the main reason for the inferior performance of the existing hard-constraining technique compared to both the vanilla PINN and the newly proposed method of hard-constraining is its significantly longer training time.
Concretely, with no Fourier feature embeddings, the vanilla PINN takes (on average) $13.5\mathrm{ms}$ to complete a single training iteration, the existing hard-constraining method $63.1\mathrm{ms}$, and the newly proposed approach of hard-constraining Neumann boundary conditions via a single Fourier feature mapping takes $14.6\mathrm{ms}$.
When training all models for~$10^6$ iterations (instead of fixing the training time as in \figref{fig:acc_bars}), it can be observed that the existing hard-constraining technique mostly performs better than the best vanilla PINN, but still worse than the newly proposed method, cf.\ \figref{fig:acc_bars_allit} in appendix~\ref{sc:details_experiments}.
This shows that hard-constraining boundary conditions indeed improves the performance of PINNs due to the reduction of the number of training task, which is consistent with the analyses from~\citet{hPINN21,Zeinhofer24} in the case of Dirichlet conditions.
The newly proposed approach offers a computationally efficient way of benefiting from this effect in the case of Neumann boundary conditions.

The results further show that choosing a suitable Fourier embedding strategy is crucial to obtain accurate results -- sufficiently large Fourier embeddings are needed to approximate high frequency behavior and no Fourier embeddings should be used for low frequency behavior. 
While this can be observed for all hard-constraining techniques, \figref{fig:acc_bars} indicates that it is particularly important for the newly proposed approach.
Some guidelines on how to choose the Fourier embedding's parameters (i.e., its size and the distribution from which to sample the random frequencies) are outlined in~\citet{Wang21}.

\section{Conclusion}\label{sc:conc}

We have presented a new approach utilizing specific Fourier feature embeddings to hard-constrain Neumann boundary conditions into PINNs.
Compared to existing hard-constraining techniques for Neumann boundary conditions, the new method is easier to implement and more computationally efficient.
In addition, it demonstrates superior performance in solving a diffusion problem, particularly excelling in high frequency and multiscale settings.
Moreover, although we have focused on a simple forward diffusion problem in this study, the method's versatility allows for its application across various contexts including inverse problems and operator learning tasks.
The new approach is expected to serve as a valuable tool in representing multiscale processes through PINNs, although further exploration is needed to fully understand its capabilities.

\subsubsection*{Acknowledgments}
This work was supported by 
the Fraunhofer Internal Programs under Grant No.\ PREPARE~40-08394.

\subsubsection*{Data availability statement}
The source code used for the experiments is publicly available at
\begin{center}
	\url{https://github.com/FhG-IISB-MKI/HC_Neumann_Experiments}
\end{center}

%

\bibliography{iclr2025_conference}
\bibliographystyle{iclr2025_conference}

\appendix

\section{Theoretical background on the one-dimensional diffusion problem}\label{sc:heat}

In this appendix, we provide the analytical formula of the solution of the diffusion problem stated in~\eqrefsys{eq:heat}{eq:initial_heat}.
Applying basic separation-of-variables and superposition techniques~\citep{Evans} yields that the unique solution~$u$ of the system is of the form
\begin{equation}\label{eq:heat_solformula}
	u(x,t)=\frac{a_0}2 + \sum_{j=1}^\infty a_j\,\exp\left(-D\pi^2j^2t\right)\,\cos\left(\pi jx\right),\qquad x,t\in[0,1],
\end{equation}
where $(a_j)_{j\in\mathbb N_0}\subset\R$ are the Fourier coefficients of the initial condition~$g$:
\begin{equation}\label{eq:Fourier_coeff}
	a_j=2\int_0^1g(x)\,\cos\left(\pi jx\right)\,dx,\qquad j\in\mathbb N_0.
\end{equation}
More details on how the limit in~\eqref{eq:heat_solformula} can be interpreted under which assumptions on~$g$ can be found in~\citep{Evans,Pinsky}.
In the case of sufficiently smooth initial data~$g$ satisfying the vanishing Neumann boundary conditions (\eqref{eq:Neumann_heat}), i.e., $g'(0)=0=g'(1)$, the infinite sum in~\eqref{eq:heat_solformula} converges uniformly on $[0,1]^2$~\citep{Pinsky}.

In the case where~$g$ consists of a single spatial frequency, all Fourier coefficients~$a_j$ except of one vanish.
Concretely, if $g(x)=\cos(\pi nx)$ for some $n\in\mathbb N$, evaluating the integrals in~\eqref{eq:Fourier_coeff} yields $a_n=1$ and $a_j=0$ for $j\neq n$. 
This is because $(\cos(\pi m\cdot))_{m\in\mathbb N_0}$ is an orthogonal system of $L^2([0,1])$~\citep{Pinsky}.
In this case, the solution of the diffusion problem given by \eqrefsys{eq:heat}{eq:initial_heat} is therefore of the form
\begin{equation}
	u(x,t)=\exp\left(-D\pi^2n^2t\right)\,\cos\left(\pi nx\right),\qquad x,t\in[0,1].
\end{equation} 
This formula gives the explicit solutions in the low frequency and high frequency cases described in \tabref{table:initials_heat} and, due to the linearity of the diffusion problem, also in the multiscale case.

In the case of simple initial data~$g$, the Fourier coefficients can be computed explicitly.
For instance, if~$g$ is a polynomial, this is possible by integrating by parts in the integrals from~\eqref{eq:Fourier_coeff}.
In this way, one can obtain the explicit solution formula for the two polynomials from \tabref{table:initials_heat}. 
Notice, however, that in this case, the sum in~\eqref{eq:heat_solformula} is indeed infinite.
For the numerics, we thus have to truncate the sum, i.e., we approximate~\eqref{eq:heat_solformula} by
\begin{equation}
	u(x,t)\approx\frac{a_0}2 + \sum_{j=1}^N a_j\,\exp\left(-D\pi^2j^2t\right)\,\cos\left(\pi jx\right),\qquad x,t\in[0,1],
\end{equation}
for some sufficiently large $N\in\mathbb N$. 
For the numerical experiments conducted in \secref{sc:experiments}, we used $N=200$ terms.

\section{Methodology -- More general cases}\label{sc:metho_general}

In \secref{sc:metho}, the formulae for hard-constraining Neumann boundary conditions are restricted to the case of one spatial variable~$x$ in the domain $[0,1]$ and boundary conditions at both parts of the boundary.
We now show that the same idea can be used to hard-constrain Neumann boundary conditions on general intervals, only on one part of the boundary, and in higher dimensional spatial domains.

In all of the below, we again consider the case where the solution has an additional time dependency, although this is not necessary.

\subsection{General intervals}

We first consider Neumann boundary conditions on a general interval $[\alpha,\beta]$, i.e.,
\begin{equation}\label{eq:Neumann_heat_generalint}
	\partial_xu(\alpha,t)=A,\qquad \partial_xu(\beta,t)=B,
\end{equation}
for a sought function $u\colon[\alpha,\beta]\times[0,T]\to\R$ and prescribed $\alpha<\beta$, $T>0$, and $A,B\in\R$. 
\Eqref{eq:Neumann_heat_generalint} can be hard-constrained by suitably rescaling the functions from \secref{sc:metho}.
Concretely, the Fourier embeddings defined in \eqref{eq:FourierEmb_NewSingle} and \eqref{eq:FourierEmb_NewMult} need to be replaced by
\begin{equation}\label{eq:FourierEmb_NewSingle_generalint}
	[\alpha,\beta]\ni x\mapsto\FourierEmb_1(x)=\cos\left(\pi\frac{x-\alpha}{\beta-\alpha}\right)
\end{equation}
and
\begin{equation}\label{eq:FourierEmb_NewMult_generalint}
	[\alpha,\beta]\ni x\mapsto\FourierEmb_{1,b_2,\ldots,b_{n}}(x)=\left(\cos\left(\pi\frac{x-\alpha}{\beta-\alpha}\right), \cos\left(\pi b_2\frac{x-\alpha}{\beta-\alpha}\right),\ldots,\cos\left(\pi b_n\frac{x-\alpha}{\beta-\alpha}\right)\right),
\end{equation}
respectively, where $b_2,\ldots,b_n\in\mathbb Z$ are again (randomly sampled) integer frequencies.
The output transformation from \eqref{eq:utrafo_our_nonvanishing} adjusting for non-vanishing Neumann boundary data becomes
\begin{equation}\label{eq:utrafo_our_nonvanishing_generalint}
	u^\trafo(x,t)=u^\NN\big(\FourierEmb(x),t\big) + (x-\alpha)(\beta-x)^2A+(x-\alpha)^2(x-\beta)B.
\end{equation} 
where $\FourierEmb$ can be $\FourierEmb_1$ or $\FourierEmb_{1,b_2,\ldots,b_{n}}$.
It is straight-forward to verify via the chain rule that $u^\trafo$ defined by \eqref{eq:utrafo_our_nonvanishing_generalint} indeed satisfies \eqref{eq:Neumann_heat_generalint} for any smooth representation model~$u^\NN$.

\subsection{One-sided Neumann boundary condition}\label{ssc:metho_oneside}

We next consider the situation where only a Neumann boundary condition at one side of the boundary is hard-constrained, i.e.,
\begin{equation}\label{eq:Neumann_heat_oneside}
	\partial_xu(\alpha,t)=A,
\end{equation}
for a sought function $u\colon[\alpha,\beta]\times[0,T]\to\R$ and prescribed $\alpha<\beta$, $T>0$, and $A\in\R$. 
In \eqref{eq:Neumann_heat_oneside}, the Neumann boundary condition is imposed at the left side of the spatial interval~$[\alpha,\beta]$; straightforward modifications of the formulae below allow to treat the right boundary in the same way.
To hard-constrain \eqref{eq:Neumann_heat_oneside}, we use the feature transformation
\begin{equation}\label{eq:FourierEmb_NewSingle_oneside}
	[\alpha,\beta]\ni x\mapsto\FourierEmb_1(x)=\cos\left(\frac\pi2\frac{x-\alpha}{\beta-\alpha}\right).
\end{equation}
This mapping is a suitable substitute for \eqref{eq:FourierEmb_NewSingle} because $\FourierEmb_1'(\alpha)=0$ and $\FourierEmb_1'(x)\neq0$ for $\alpha<x\leq \beta$.  The latter is needed for the transformed model to be able to learn general derivative values on $]\alpha,\beta]$, in particular, at the right boundary $x=\beta$.
To better capture high-frequency phenomena, the above transformation can be extended to 
\begin{equation}\label{eq:FourierEmb_NewMult_oneside}
	[\alpha,\beta]\ni x\mapsto\FourierEmb_{1,b_2,\ldots,b_{n}}(x)=\left(\cos\left(\frac\pi2\frac{x-\alpha}{\beta-\alpha}\right), \cos\left(\frac\pi2 b_2\frac{x-\alpha}{\beta-\alpha}\right),\ldots,\cos\left(\frac\pi2 b_n\frac{x-\alpha}{\beta-\alpha}\right)\right).
\end{equation}
In this case, the (randomly sampled) frequencies $b_2,\ldots,b_n$  can be arbitrary real numbers (not necessarily integers).
The key property $\frac\partial{\partial x}\FourierEmb_{1,b_2,\ldots,b_{n}}(\alpha)=0$ is ensured by using only Cosine mappings and no Sines.
The overall transformation to hard-constrain \eqref{eq:Neumann_heat_oneside} for a arbitrary smooth representation model~$u^\NN$ takes on the form
\begin{equation}\label{eq:utrafo_our_nonvanishing_oneisde}
	u^\trafo(x,t)=u^\NN\big(\FourierEmb(x),t\big) + (x-\alpha)A,
\end{equation} 
where $\FourierEmb$ can be defined by \eqref{eq:FourierEmb_NewSingle_oneside} or \eqref{eq:FourierEmb_NewMult_oneside}.

\subsection{Neumann boundary condition in higher dimensions}\label{ssc:metho_higherdim}

Lastly, we show how the newly proposed method to hard-constrain Neumann boundary conditions can be applied in the case where the spatial domain is a hyperrectangle $[\alpha_1,\beta_1]\times\ldots\times[\alpha_d,\beta_d]$ of dimension $d\in\mathbb N$ for prescribed $\alpha=(\alpha_1,\ldots,\alpha_d),b=(\beta_1,\ldots,\beta_d)\in\R^d$ with $\alpha_i<\beta_i$ for $i=1,\ldots,d$.
We consider the case where Neumann boundary conditions are imposed at all parts of the boundary, i.e.,
\begin{align}
	\partial_{x_i}u(x_1,\ldots,x_{i-1},\alpha_i,x_{i+1},\ldots,x_d,t)&=A_i,\ &&i=1,\ldots,d,\label{eq:Neumann_heat_higherdim_a}\\
	\partial_{x_i}u(x_1,\ldots,x_{i-1},\beta_i,x_{i+1},\ldots,x_d,t)&=B_i,\ &&i=1,\ldots,d,\label{eq:Neumann_heat_higherdim_b}
\end{align}
where $A_1,\ldots,A_d,B_1,\ldots,B_d\in\R$ are given and $u\colon[\alpha_1,\beta_1]\times\ldots\times[\alpha_d,\beta_d]\times[0,T]\to\R$ denotes the sought solution.
To hard-constrain \eqrefsys{eq:Neumann_heat_higherdim_a}{eq:Neumann_heat_higherdim_b}, the transformations from \eqrefsys{eq:FourierEmb_NewSingle_generalint}{eq:FourierEmb_NewMult_generalint} need to be applied to all input dimensions.
Concretely, let
\begin{equation}\label{eq:FourierEmb_NewSingle_higherdim}
	[\alpha_1,\beta_1]\times\ldots\times[\alpha_d,\beta_d]\ni x\mapsto\FourierEmb_1(x)=\left(\cos\left(\pi\frac{x_1-\alpha_1}{\beta_1-\alpha_1}\right),\ldots,\cos\left(\pi\frac{x_d-\alpha_d}{\beta_d-\alpha_d}\right)\right)
\end{equation}
and
\begin{multline}\label{eq:FourierEmb_NewMult_higherdim}
	[\alpha_1,\beta_1]\times\ldots\times[\alpha_d,\beta_d]\ni x\mapsto\FourierEmb_{1, b_2,\ldots, b_{n}}(x)=\\=\left(\cos\left(\pi\frac{x-\alpha}{\beta-\alpha}\right), \cos\left(\pi  b_2\frac{x-\alpha}{\beta-\alpha}\right),\ldots,\cos\left(\pi  b_n\frac{x-\alpha}{\beta-\alpha}\right)\right).
\end{multline}
In the latter expression, the frequencies $ b_2,\ldots, b_{n}$ are again required to be integers and all vector-operations are meant in a component-wise way, so that the output of $\FourierEmb_{1, b_2,\ldots, b_{n}}$ is of dimension $n\cdot d$.
The overall transformation to hard-constrain \eqrefsys{eq:Neumann_heat_higherdim_a}{eq:Neumann_heat_higherdim_b} is
\begin{equation}\label{eq:utrafo_our_nonvanishing_higherdim}
	u^\trafo(x,t)=u^\NN\big(\FourierEmb(x),t\big)+\sum_{i=1}^d\Bigg(A_i(x_i-\alpha_i)\prod_{\substack{j=1,\\j\neq i}}^d(\beta_j-x_j)^2 + B_i(x_i-\beta_i)\prod_{\substack{j=1,\\j\neq i}}^d(x_j-\alpha_j)^2\Bigg)
\end{equation}
for $x=(x_1,\ldots,x_n)\in[\alpha_1,\beta_1]\times\ldots\times[\alpha_d,\beta_d]$ and $0\leq t\leq T$,
where $\FourierEmb$ can be defined by \eqref{eq:FourierEmb_NewSingle_higherdim} or \eqref{eq:FourierEmb_NewMult_higherdim} and $u^\NN$ is an arbitrary smooth representation model.

In order to not hard-constrain Neumann boundary conditions for certain spatial input variables, the feature transformations from \eqrefsys{eq:FourierEmb_NewSingle_higherdim}{eq:FourierEmb_NewMult_higherdim} can simply be modified so that they leave the corresponding variables unchanged.
Hard-constraining Neumann boundary conditions only at one part of the boundary in certain spatial input dimension can be achieved by modifications similar to those presented in \secref{ssc:metho_oneside}.

\section{More details on the experiments}\label{sc:details_experiments}

The aim of this appendix is to provide further details on the experiments conducted and mentioned in \secref{sc:experiments}.

Firstly, we specify the hyperparameters used for all numerical experiments in this study.
As mentioned in \secref{sc:experiments}, these hyperparameters are chosen similarly to those used in~\citet{Wang21}.
The representation model is always a fully connected neural network with~$3$ hidden layers consisting of~$100$ neurons each. 
The activation function is $\tanh$.
The models are trained with a fixed learning rate of~$10^{-4}$ using the Adam optimizer.
The numbers of collocation points used to compute the loss~$\L$ are $N_\PDE=2\cdot10^4$, $N_\IC=500$, and $N_\BC=10^3$.
In all settings, the accuracy is computed for the model with the lowest (training) loss encountered during training.
The code is written in DeepXDE~\citep{DeepXDE21} with backend PyTorch~\citep{PyTorch}.
All experiments were carried out on an NVIDIA Quadro RTX 5000 (16GB RAM).

Secondly, we provide more details on the experiments visualized in \figref{fig:acc_bars}.
Concretely, raw data for the accuracies (relative $L^2$-error w.r.t.\ the analytical reference solution) of all models from \figref{fig:acc_bars} are given in \tabref{table:experiments}.

\begin{table}[h]
	\caption{Accuracies for the experiments visualized in \figref{fig:acc_bars} in the settings given by in \tabref{table:initials_heat}. In each setting, the data corresponding to the reference method (which is always chosen as the vanilla PINN with the Fourier embedding strategy leading to the best accuracy) is shown in italic font. The data corresponding to the best accuracy is highlighted in bold font.}
	\label{table:experiments}
	\begin{center}
		\begin{tabular}{|l||l|l|l|l|l|} 
			\hline
			 & \multirow{2}{4.5em}{Low frequency} & \multirow{2}{4.5em}{High frequency}& \multirow{2}{4.5em}{Multiscale} & \multirow{2}{4.5em}{Polynom 3rd order} & \multirow{2}{4.5em}{Polynom 4th order}\\ & & & & & \\\hline\hline
			\makecell[tl]{Vanilla PINN\\\phantom{T}} & $\mathit{4.14\cdot10^{-5}}$ & $1.64\cdot10^{-1}$ & $3.80\cdot10^{-2}$ & $\mathit{1.30\cdot10^{-4}}$ & $\mathit{1.61\cdot10^{-4}}$ \\\hline
			\makecell[tl]{Vanilla PINN\\ w/ Fourier~$20$} & $9.54\cdot10^{-5}$ & $5.49\cdot10^{-3}$ & $4.22\cdot10^{-4}$ & $3.85\cdot10^{-4}$ & $5.66\cdot10^{-4}$ \\\hline
			\makecell[tl]{Vanilla PINN\\ w/ Fourier~$50$} & $8.34\cdot10^{-4}$ & $\mathit{3.69\cdot10^{-3}}$ & $\mathit{2.40\cdot10^{-4}}$ & $3.66\cdot10^{-3}$ & $4.26\cdot10^{-3}$ \\\hline
			\makecell[tl]{Existing HC\\\phantom{T}} & $6.58\cdot10^{-5}$ & $4.50\cdot10^{-1}$ & $6.57\cdot10^{-2}$ & $1.40\cdot10^{-4}$ & $3.39\cdot10^{-4}$ \\\hline
			\makecell[tl]{Existing HC\\ w/ Fourier~$20$} & $7.33\cdot10^{-4}$ & $8.08\cdot10^{-3}$ & $9.86\cdot10^{-4}$ & $4.13\cdot10^{-3}$ & $3.99\cdot10^{-3}$ \\\hline
			\makecell[tl]{Existing HC\\ w/ Fourier~$50$} & $2.80\cdot10^{-2}$ & $8.42\cdot10^{-3}$ & $3.34\cdot10^{-4}$ & $1.22\cdot10^{-1}$ & $1.92\cdot10^{-1}$ \\\hline
			\makecell[tl]{New HC via\\ $1$~Fourier frequency} & $\mathbf{1.90\cdot10^{-5}}$ & $3.73\cdot10^{-1}$ & $1.15\cdot10^{-2}$ & $\mathbf{6.85\cdot10^{-5}}$ & $\mathbf{1.20\cdot10^{-4}}$ \\\hline
			\makecell[tl]{New HC via\\ $20$~Fourier frequencies} & $8.01\cdot10^{-4}$ & $1.61\cdot10^{-4}$ & $1.18\cdot10^{-4}$ & $3.13\cdot10^{-3}$ & $3.55\cdot10^{-3}$ \\\hline
			\makecell[tl]{New HC via\\ $50$~Fourier frequencies} & $9.43\cdot10^{-3}$ & $\mathbf{1.30\cdot10^{-4}}$ & $\mathbf{6.05\cdot10^{-5}}$ & $1.55\cdot10^{-1}$ & $2.02\cdot10^{-1}$ \\\hline
		\end{tabular}
	\end{center}
\end{table}

\begin{figure}[h]
	\begin{center}
		\includegraphics[width=\columnwidth]{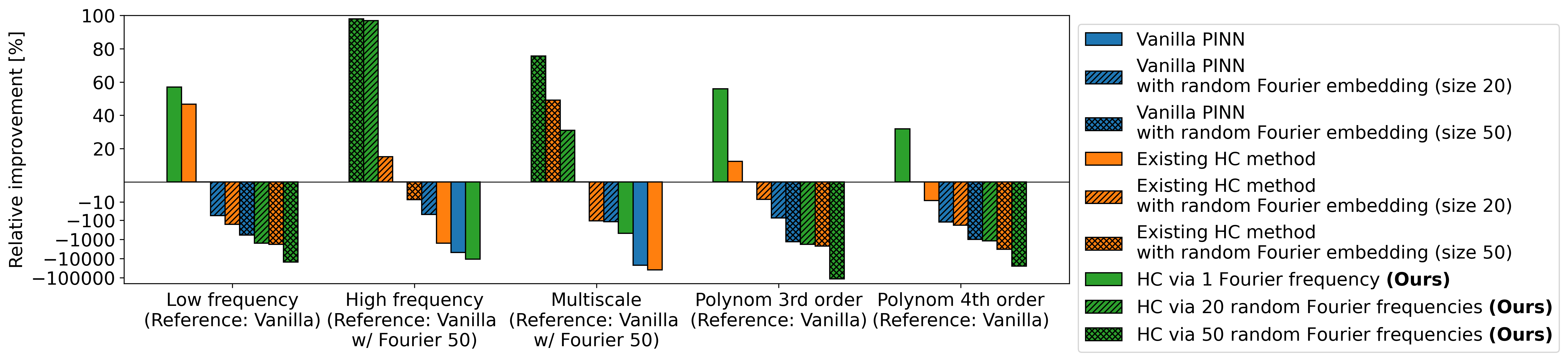}
	\end{center}
	\caption{Relative improvements of the accuracies of PINNs with different hard-constraining (HC) and Fourier embedding strategies w.r.t.\ a reference PINN in the settings described in \tabref{table:initials_heat}. Different to \figref{fig:acc_bars}, all models are trained for the same number of iterations ($10^6$ training iterations) here.}
	\label{fig:acc_bars_allit}
\end{figure}

\begin{table}[h]
	\caption{Accuracies for the experiments visualized in \figref{fig:acc_bars_allit} in the settings given by in \tabref{table:initials_heat}. Different to \tabref{table:experiments}, all models have been trained for the same number of iterations ($10^6$ training iterations) here.  In each setting, the data corresponding to the reference method (which is always chosen as the vanilla PINN with the Fourier embedding strategy leading to the best accuracy) is shown in italic font. The data corresponding to the best accuracy is highlighted in bold font.}
	\label{table:experiments_allit}
	\begin{center}
		\begin{tabular}{|l||l|l|l|l|l|} 
			\hline
			& \multirow{2}{4.5em}{Low frequency} & \multirow{2}{4.5em}{High frequency}& \multirow{2}{4.5em}{Multiscale} & \multirow{2}{4.5em}{Polynom 3rd order} & \multirow{2}{4.5em}{Polynom 4th order}\\ & & & & & \\\hline\hline
			\makecell[tl]{Vanilla PINN\\\phantom{T}} & $\mathit{4.14\cdot10^{-5}}$ & $1.64\cdot10^{-1}$ & $3.80\cdot10^{-2}$ & $\mathit{1.30\cdot10^{-4}}$ & $\mathit{1.61\cdot10^{-4}}$ \\\hline
			\makecell[tl]{Vanilla PINN\\ w/ Fourier~$20$} & $6.51\cdot10^{-5}$ & $5.00\cdot10^{-3}$ & $3.74\cdot10^{-4}$ & $2.25\cdot10^{-4}$ & $3.57\cdot10^{-4}$ \\\hline
			\makecell[tl]{Vanilla PINN\\ w/ Fourier~$50$} & $2.82\cdot10^{-4}$ & $\mathit{3.37\cdot10^{-3}}$ & $\mathit{1.72\cdot10^{-4}}$ & $1.83\cdot10^{-3}$ & $1.75\cdot10^{-3}$ \\\hline
			\makecell[tl]{Existing HC\\\phantom{T}} & $2.21\cdot10^{-5}$ & $5.60\cdot10^{-2}$ & $6.59\cdot10^{-2}$ & $1.14\cdot10^{-4}$ & $1.75\cdot10^{-4}$ \\\hline
			\makecell[tl]{Existing HC\\ w/ Fourier~$20$} & $1.09\cdot10^{-4}$ & $2.86\cdot10^{-3}$ & $3.54\cdot10^{-4}$ & $1.39\cdot10^{-4}$ & $4.55\cdot10^{-4}$ \\\hline
			\makecell[tl]{Existing HC\\ w/ Fourier~$50$} & $7.91\cdot10^{-4}$ & $3.62\cdot10^{-3}$ & $8.75\cdot10^{-5}$ & $3.04\cdot10^{-3}$ & $5.44\cdot10^{-3}$ \\\hline
			\makecell[tl]{New HC via\\ $1$~Fourier frequency} & $\mathbf{1.78\cdot10^{-5}}$ & $3.64\cdot10^{-1}$ & $1.01\cdot10^{-3}$ & $\mathbf{5.71\cdot10^{-5}}$ & $\mathbf{1.10\cdot10^{-4}}$ \\\hline
			\makecell[tl]{New HC via\\ $20$~Fourier frequencies} & $6.81\cdot10^{-4}$ & $1.06\cdot10^{-4}$ & $1.19\cdot10^{-4}$ & $2.47\cdot10^{-3}$ & $2.05\cdot10^{-3}$ \\\hline
			\makecell[tl]{New HC via\\ $50$~Fourier frequencies} & $6.28\cdot10^{-3}$ & $\mathbf{7.23\cdot10^{-5}}$ & $\mathbf{4.18\cdot10^{-5}}$ & $1.47\cdot10^{-1}$ & $4.00\cdot10^{-2}$ \\\hline
		\end{tabular}
	\end{center}
\end{table}

Recall that for the simulations visualized in \figref{fig:acc_bars}, the training time of all models is fixed to ensure a fair comparison across the different methods.
As discussed in \secref{sc:experiments}, this is the main reason for the consistently inferior performance of the existing hard-constraining method because of the longer time it takes to train.
To underpin this statement, all models from \secref{sc:experiments} have also been trained for the same number of $10^6$ iterations.
The results of these modified experiments are given below.
Concretely, \figref{fig:acc_bars_allit} and \tabref{table:experiments_allit} are the analogues of \figref{fig:acc_bars} and \tabref{table:experiments}, respectively, in the case where all models are trained for a fixed number of iterations.
As discussed in \secref{sc:experiments}, it can be seen that the models based on the existing hard-constraining technique indeed perform better than the vanilla PINN in most of the cases when trained for the same number of iterations.
Nonetheless, in all of the settings described in \tabref{table:initials_heat}, the best performances can always be observed when using the newly proposed method of hard-constraining the Neumann boundary conditions via Fourier embeddings.

\end{document}

%% file: math_commands.tex
\usepackage{amsmath,amsfonts,bm}

\def\figref#1{Figure~\ref{#1}}

\def\tabref#1{Table~\ref{#1}}
\def\secref#1{section~\ref{#1}}

\def\eqref#1{equation~\ref{#1}}
\def\Eqref#1{Equation~\ref{#1}}

\def\eqrefsys#1#2{equations~\ref{#1}--\ref{#2}}

\def\1{\bm{1}}

\DeclareMathAlphabet{\mathsfit}{\encodingdefault}{\sfdefault}{m}{sl}
\SetMathAlphabet{\mathsfit}{bold}{\encodingdefault}{\sfdefault}{bx}{n}

\newcommand{\R}{\mathbb{R}}

